%% file: SCIA_master.tex
\documentclass[runningheads]{llncs}
\usepackage[T1]{fontenc}
\usepackage{graphicx}
\usepackage{hyperref}
\usepackage{color}

\usepackage{times}
\usepackage{epsfig}
\usepackage{array,multirow,graphicx}
\usepackage{amsmath}
\usepackage{amssymb}
\usepackage{booktabs}
\usepackage{stfloats}
\usepackage{tabularx}
\usepackage{subcaption}
\usepackage{wrapfig}
\usepackage[draft,footnote,nomargin]{fixme}
\usepackage{ulem}
\usepackage{listings}
\usepackage{tikz}
\usepackage[T1]{fontenc}

\newcommand{\figref}[1]{Figure~\ref{fig:#1}}
\newcommand{\tableref}[1]{Table~\ref{tab:#1}}

\begin{document}
%
\title{BrackishMOT: The Brackish Multi-Object Tracking Dataset}
%
%


\author{Malte Pedersen\inst{1,2,*}, Daniel Lehotský\inst{1}, Ivan Nikolov\inst{1,2}, Thomas B. Moeslund\inst{1,2}}
\institute{Visual Analysis and Perception Lab, Aalborg University, Aalborg, Denmark \and Pioneer Center for AI, Denmark\\
* \email{mape@create.aau.dk}}


\authorrunning{M. Pedersen et al.}

%
\maketitle              
\begin{abstract}
    There exist no publicly available annotated underwater multi-object tracking (MOT) datasets captured in turbid environments. To remedy this we propose the BrackishMOT dataset with focus on tracking schools of small fish, which is a notoriously difficult MOT task. BrackishMOT consists of 98 sequences captured in the wild. Alongside the novel dataset, we present baseline results by training a state-of-the-art tracker.
    Additionally, we propose a framework for creating synthetic sequences in order to expand the dataset. The framework consists of animated fish models and realistic underwater environments.
    We analyse the effects of including synthetic data during training and show that a combination of real and synthetic underwater training data can enhance tracking performance.
    \textit{Links to code and data can be found at \url{https://www.vap.aau.dk/brackishmot}}.


\end{abstract}

\section{Introduction}\label{sec:intro}
\input{sections/introduction}

\section{Related Work}\label{sec:related}
\input{sections/related_work}

\section{The BrackishMOT Dataset}\label{sec:brackish}
\input{sections/brackishMOT}
\input{sections/split}

\section{Synthetic Data Framework}\label{sec:synth}
\input{sections/syntheticData}

\section{Experiments}\label{sec:baseline}
\input{sections/results_new}

\section{Conclusion}
\input{sections/conclusion}

\subsubsection{Acknowledgements}
This work has been funded by the Independent Research Fund Denmark under the case number 9131-00128B.

{
    \small
    \bibliographystyle{splncs04}
    \bibliography{bibliography}
}
%
%
%
\end{document}

%% file: sections/introduction.tex
Humans have relied on the oceans as a steady food-source for millennia, but the marine ecosystems are now rapidly deteriorating due to human impact.
This is especially a problem for coastal societies across the globe that rely on fish as their main food-source and for biodiversity in general.
The severity is underlined by the fact that the United Nations included  \textit{Life Below Water} as the fourteenth Sustainable Development Goal (UN SDG \#14)~\cite{ungoals}.
The increase in attention to monitoring the condition of the ocean has entailed pressure on marine researchers to gather data at an unprecedented pace.
However, traditional marine data-gathering methods are often time-consuming, intrusive, and difficult to scale as they require organisms to be caught and measured manually.
Therefore, it is critical that assistive solutions for optimizing and scaling data-gathering in marine environments are developed.

\begin{figure}[t]
    \centering
    \includegraphics[width=0.9\linewidth]{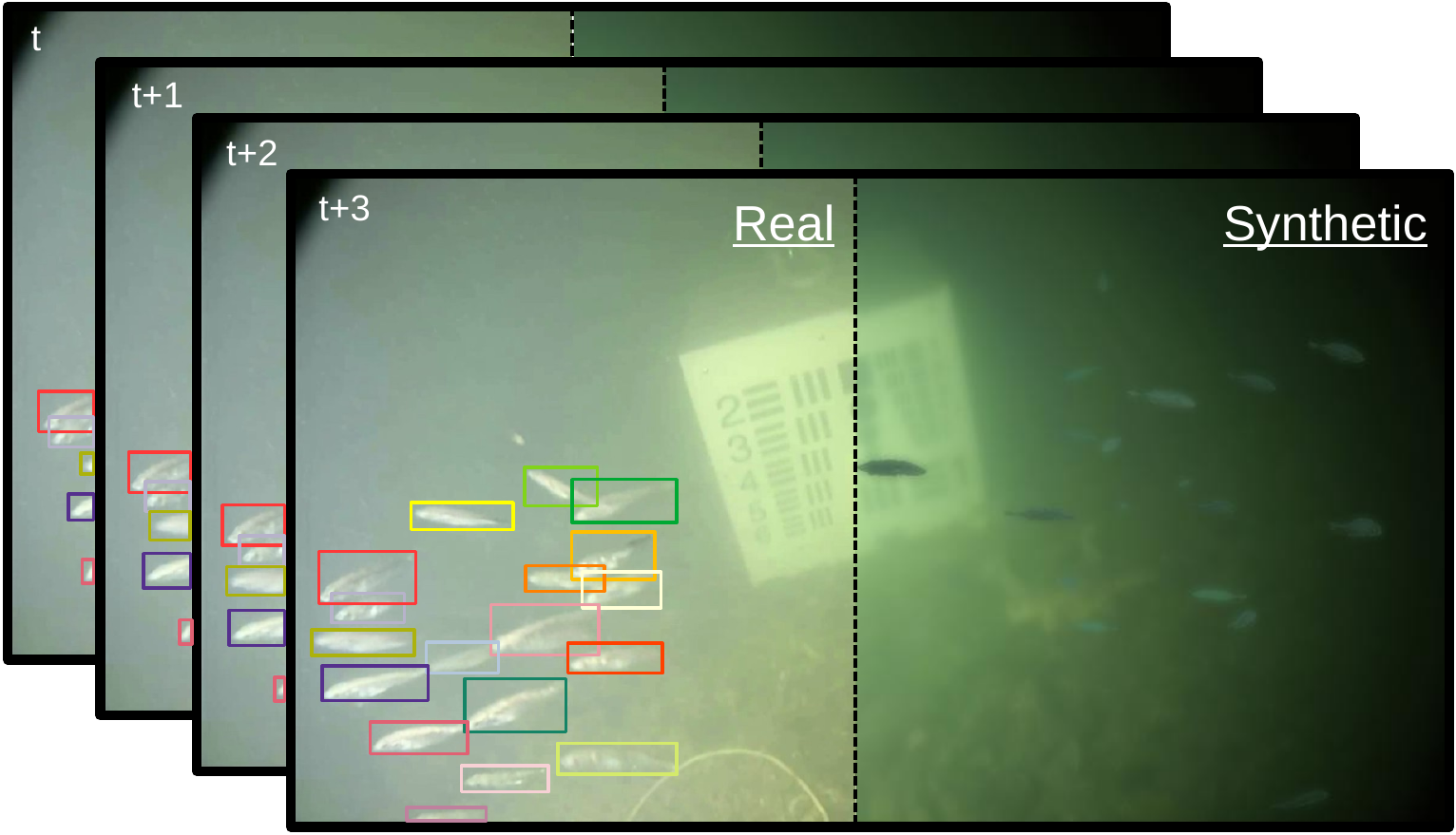}
    \caption{We present and publish a bounding box annotated underwater multi-object tracking dataset captured in the wild named BrackishMOT, together with a synthetic framework for generating more data for which we publish both data and source code.}
    \label{fig:frontpage}
\end{figure}

During the past decade, computer vision solutions have increased dramatically in performance due to the utilization of strong graphics processing units (GPU) combined with the popularisation of deep learning algorithms.
Simultaneously, underwater cameras have become significantly better and cheaper \cite{madsen2021fishing}.
This calls for marine researchers to utilize both cameras and computer vision to scale data-gathering right away, however, this is currently not feasible as there is a critical lack of marine datasets for training and evaluating marine computer vision models.

To remedy this gap we propose a novel multi-object tracking (MOT) dataset of marine organisms in the wild named BrackishMOT.
In short, MOT describes the task of obtaining the trajectories of all objects in the scene.
The trajectories are obtained by having a model detect objects spatially and associating the detections temporally.
Tracking is a core component in marine research and can be used for multiple purposes such as counting or conducting behavioral analysis.

Manually annotated datasets are critical and necessary for evaluating the performance of trackers on data from the wild.
However, they are not scalable and do not necessarily generalize well to environments that are not included in the dataset.
Therefore, we investigate how synthetic underwater sequences can be used for training multi-object trackers.
We develop a framework for creating synthetic sequences that resemble the BrackishMOT environment and analyse how key factors, namely turbidity, floating particles, and the background, affect tracker performance.
This is a critical step toward the development of new high-quality underwater synthetic datasets.
Our contributions are summarized below:

\subsubsection*{Contributions}
\begin{itemize}
    \item We present and publish BrackishMOT, a novel MOT dataset captured in brackish waters with a total of 98 sequences and six different classes. 
    \item We propose a framework for creating synthetic underwater sequences based on phenomena observed in the wild and analyse their effect on tracker performance.
    \item We analyse different training strategies for a state-of-the-art tracker using both real and synthetic data and present baseline results for BrackishMOT. 
\end{itemize}

%% file: sections/related_work.tex
The current state-of-the-art MOT algorithms like Tracktor~\cite{Bergmann_2019_ICCV}, CenterTrack~\cite{zhou2020tracking}, FairMOT~\cite{zhang2021fairmot}, and ByteTrack~\cite{Zhang2022} have all been developed for tracking terrestrial objects like pedestrians and vehicles. 
Common denominators for these types of objects are relatively predictable motion and typically strong visual cues.
However, in the underwater domain, most objects like fish are prey animals which means that they may behave erratically to avoid being tracked by predators.
Furthermore, objects of the same species often look very similar and provide weak visual features for re-identification.
In other words, the trackers need to be re-trained or fine-tuned to cope with the challenges of the underwater domain.
Modern trackers are generally based on deep learning and require large amounts of training data.
While there exist multiple terrestrial tracking datasets like KITTI~\cite{Geiger2012CVPR}, MOTChallenge~\cite{leal2015motchallenge,milan2016mot16,dendorfer2020mot20}, and UAVDT~\cite{Yu2019}, there are only a few publicly available datasets with underwater objects. 
In this section, we dive into the sparse literature on underwater datasets and trackers.

\subsection{Underwater MOT Datasets}\label{sec:datasets}
Compared to its terrestrial counterpart, the underwater MOT domain has not witnessed a noteworthy increase in novel algorithms during the past decade.
One of the reasons for this lack of algorithms dedicated to the underwater domain is the low number of publicly available annotated underwater datasets suitable for training and evaluating modern algorithms.

Underwater MOT datasets can generally be split into two categories: controlled environments such as aquariums~\cite{romero2019idtracker,pedersen20203d}, and the more challenging uncontrolled natural underwater environments which we will focus on in this paper. 
One of the earliest datasets used for tracking of fish captured in the wild was the Fish4Knowledge (F4K) dataset \cite{fisher2016fish4knowledge,Giordano2016}.
The F4K dataset was captured more than a decade ago in mostly clear tropical waters off the coast of Taiwan in very low resolution and low frame rate.
More recently, the two underwater object tracking datasets UOT32 \cite{kezebou2019underwater} and UOT100 \cite{panetta2021comprehensive} were published with annotated underwater sequences sourced from YouTube videos.
The UOT32 and UOT100 datasets provide sequences from diverse underwater environments but are focused on single object tracking.
Lastly, a high-resolution underwater MOT dataset captured off the coast of Hawai'i island named FISHTRAC \cite{mandel2023detection} was recently proposed.
However, at the time of writing only three training videos (671 frames in total) with few objects and little occlusion have been published.

The datasets captured in tropical waters only cover a tiny fraction of the diverse underwater ecosystems.
The conditions in many other areas are far less favorable with less colorful fish and more turbid water.
To advance the research in underwater MOT, it is critical to developing new datasets captured in other and more challenging environments and we see the BrackishMOT dataset as an important contribution to this field.

\subsection{Underwater Trackers}
Relatively few multi-object trackers dedicated to the underwater domain exist with most of them developed for tracking fish in controlled environments~\cite{romero2019idtracker,pedersen20203d,barreiros2021zebrafish,wang2022real}.
A common trait for trackers developed for controlled environments is the assumption of good detections and strong visual cues for re-identification.
This is generally not the case in uncontrolled environments, where the light may change, the water is turbid, the background varies, and algae may bloom on the lens~\cite{fisher2016fish4knowledge,pedersen2021no}.

To tackle these problems the team behind the F4K dataset proposed a method for detecting fish using mixture models for background subtraction and handcrafted features based on motion and color for classifying fish from other objects~\cite{fisher2016fish4knowledge}. 
Lastly, they modeled every track by feature-based covariance matrices based on representations from previous frames and associated new detections by minimizing the distance between the covariance matrices.
Another group that also worked on the F4K tracking data experimented with AlexNet~\cite{alexNet2012} and VGG-19~\cite{vgg2014} as feature extractors for appearance-based association and used a directed acyclic graph in a two-step approach by first constructing strong local tracklets followed by a tracklet-association step for finalizing the tracks~\cite{jager2017visual}.

Recently, a few groups have proposed trackers evaluated on new annotated underwater datasets.
Liu et al. proposed a multi-class tracker named RMFC~\cite{liu2021multi} utilizing YOLOv4~\cite{bochkovskiy2020yolov4} as the backbone for a detection and tracking branch running in parallel, which showed promising results.
In the work by Martija et al.~\cite{martija2021syndhn} they investigated the use of synthetic data to enhance tracker performance.
They propose to use Faster R-CNN~\cite{Ren2017} for object detection and a deep hungarian network\cite{xu2020train} for associating detections temporally using visual cues.
Unfortunately, both groups evaluate their method on private datasets, and they have not shared their code. 

The most recent work on fish tracking in the wild is the work done by Mandel et al.~\cite{mandel2023detection}.
They propose an offline tracker utilizing a greedy approach that initializes a track from the strongest detection across all frames based on a confidence score.
Detections in previous and future frames are associated with the track based on appearance and motion.
When a track has been finalized, the next track is built in the same manner, and so forth.
They evaluated their tracker with detections from YOLOv4 and RetinaNet~\cite{Lin2020}. 

Common for the aforementioned methods is a reliance on strong visual cues or predictable motion for associating detections.
This works well for scenes with few objects, in clear tropical waters, or if the objects are visually distinct.
This is a natural consequence caused by the limited datasets used in the development of the methods and it exposes the need for diverse datasets to represent the variety of underwater ecosystems.

\subsection{Synthetic Underwater Data}
A way to remedy the scarce amount of publicly available underwater datasets is to use synthetic data.
A typical approach to produce synthetic underwater data (in 2D) is by pasting cutouts of the organisms onto some background. 
Mahmood et al.\cite{Mahmood2019} used this method to place manually segmented parts of lobsters, like the body or the antennas, onto a diverse set of backgrounds sampled from the Benthoz15~\cite{bewley2015australian} dataset to generate synthetic data suitable for lobster detection in heavily occluded scenes. 
Martija et al.~\cite{martija2021syndhn} used weakly generated bounding boxes and masks to simulate the movement of fish across a background to create rough synthetic MOT data. 
And for developing an underwater litter detector Music et al.~\cite{music2020detecting} pasted various 3D shapes into real-life underwater images to create training data.

An alternative to the 2D approach is data generation from animated 3D scenes.
In 1987, Reynolds proposed the boid model for accurately simulating the behaviour of fish schools~\cite{reynolds1987flocks}.
The boid behaviour model has since been extended multiple times~\cite{stephens2003modelling,hartman2006autonomous,Podila2017}, with~\cite{ishiwaka2021foids} combining their variation of boids with a synthetic data generator to produce realistic annotated underwater data from animated sequences of fish schools.
However, to the best of our knowledge, there has been no attempt to produce synthetic underwater MOT data based on boid behavior.
In the next section, we will introduce and describe our new underwater MOT dataset followed by a description of our framework for creating synthetic underwater sequences.

%% file: sections/brackishMOT.tex
\begin{figure}[]
    \centering  
\begin{subfigure}[b]{0.32\linewidth}
    \includegraphics[width=\linewidth]{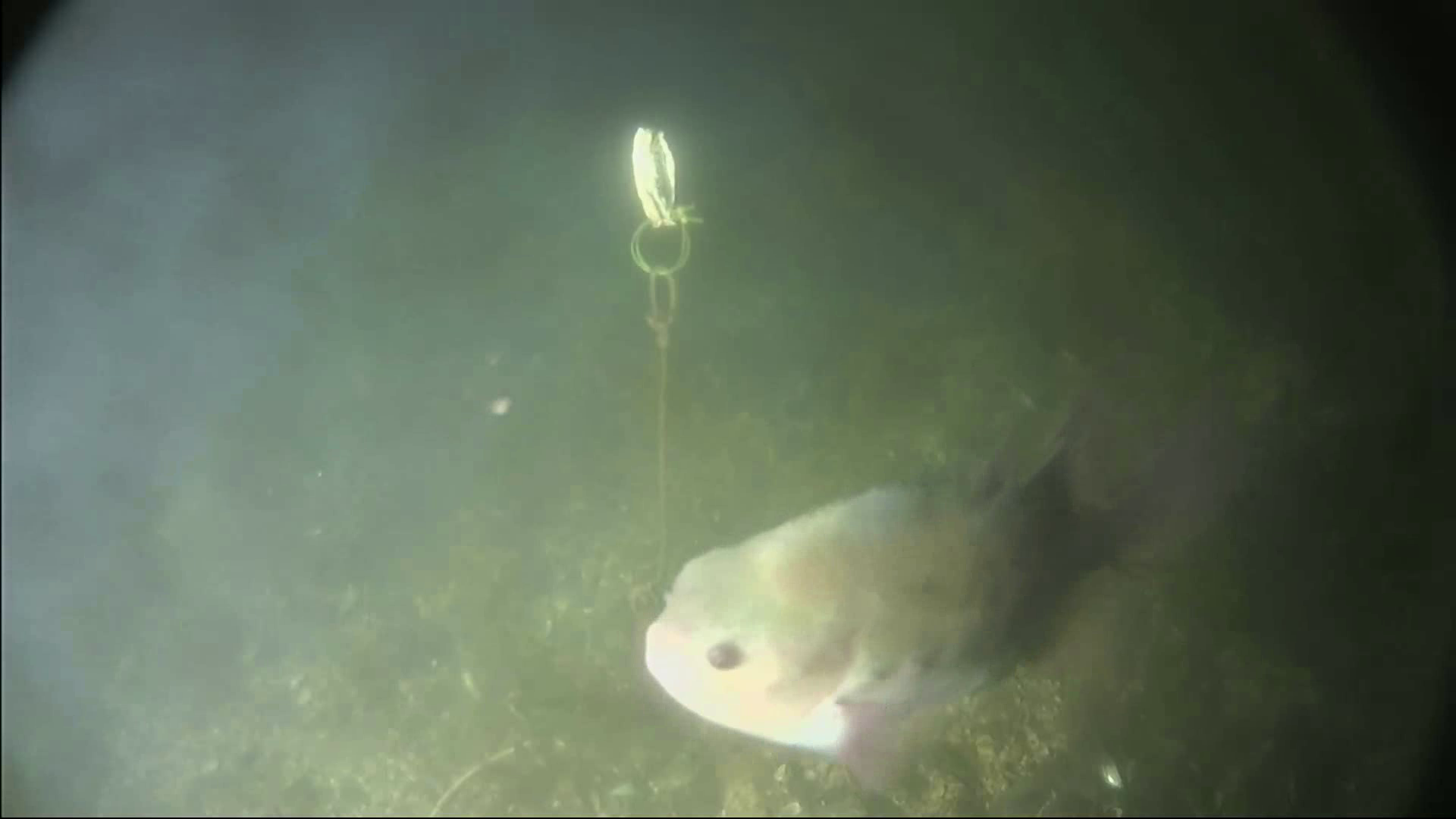}
    \caption{Fish}
    \label{fig:brackishA}
\end{subfigure}
\begin{subfigure}[b]{0.32\linewidth}
    \includegraphics[width=\linewidth]{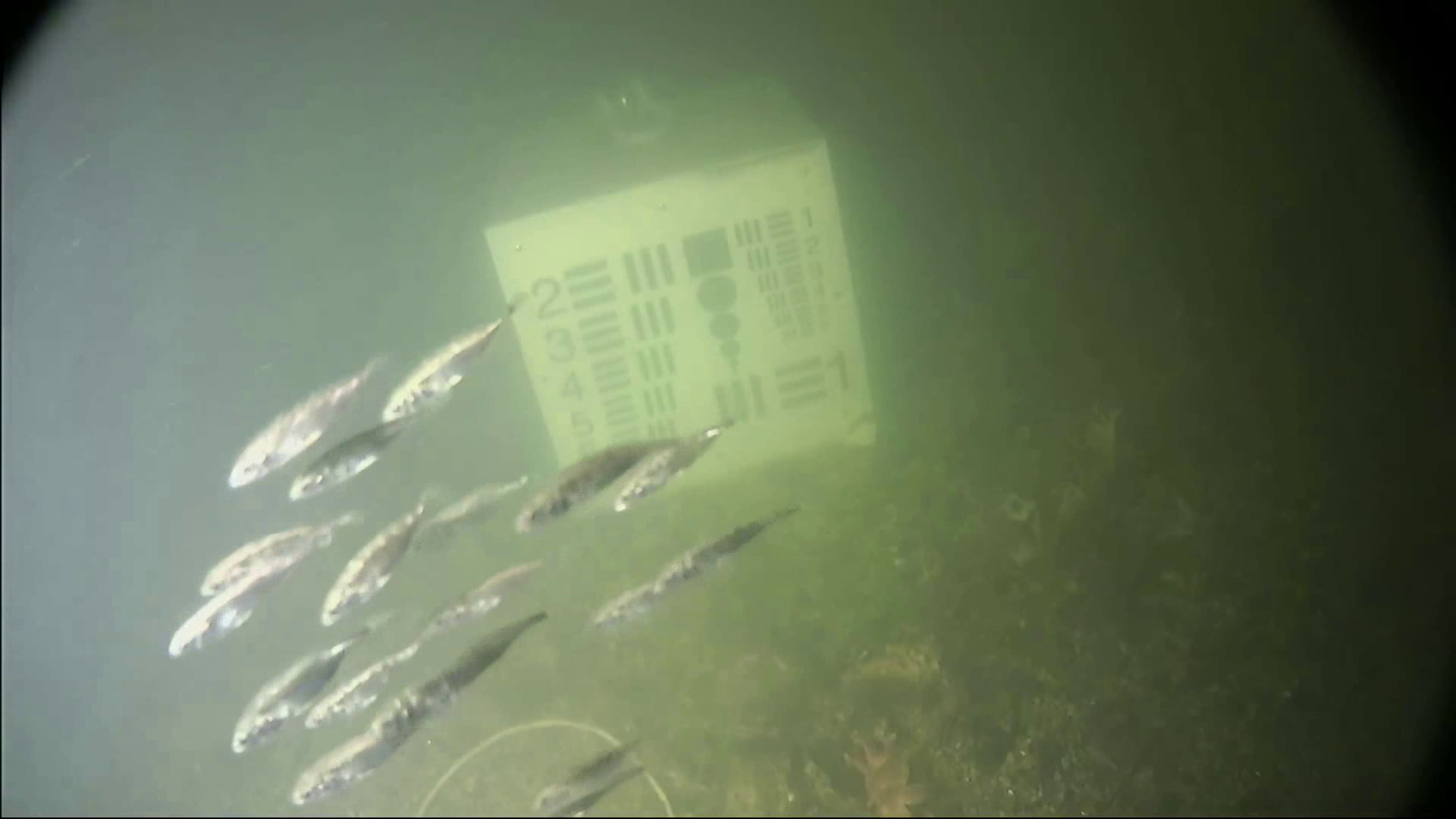}
    \caption{Small fish}
    \label{fig:brackishB}
\end{subfigure}
\begin{subfigure}[b]{0.32\linewidth}
    \includegraphics[width=\linewidth]{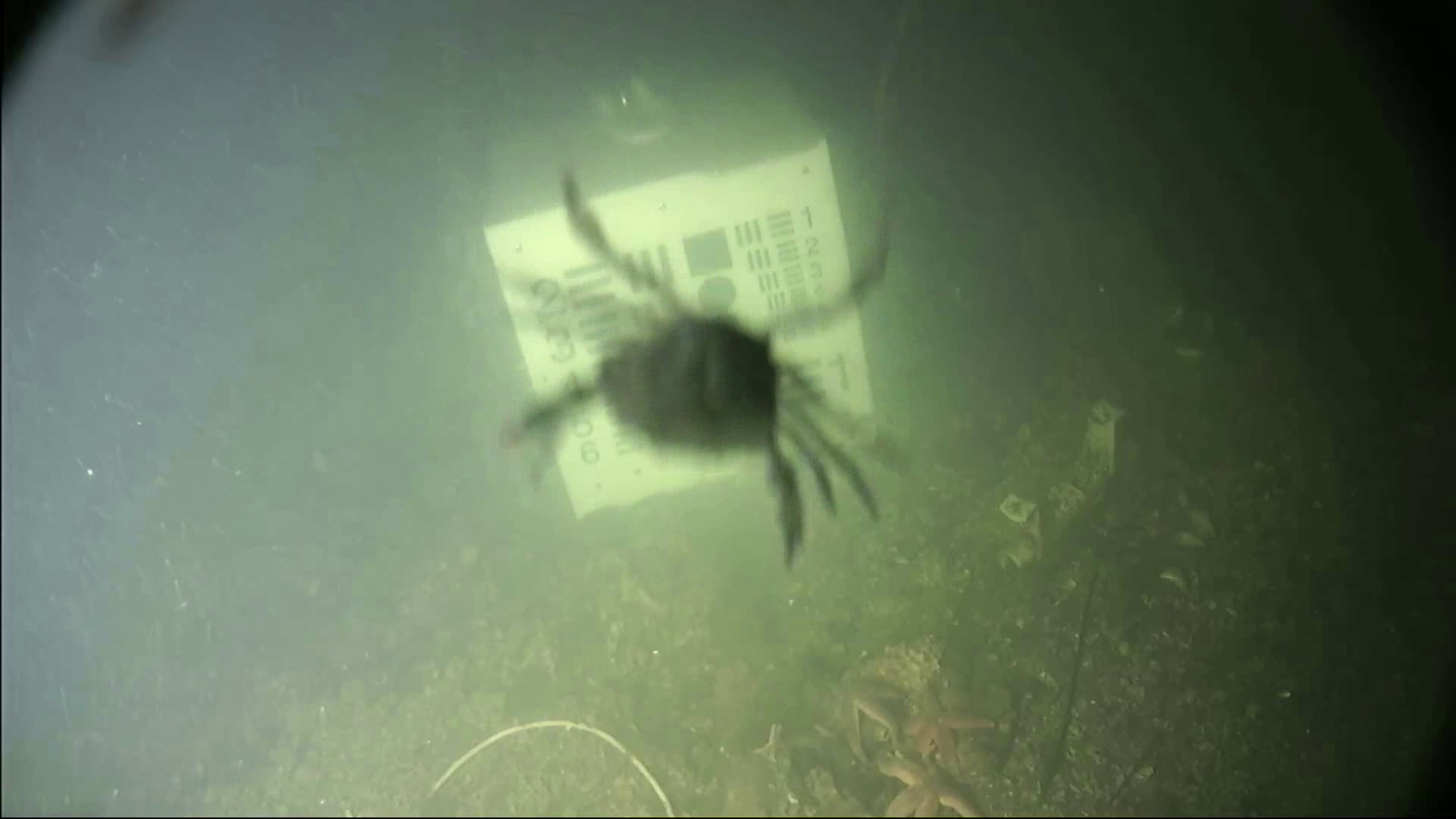}
    \caption{Crab}
    \label{fig:brackishC}
\end{subfigure}
    \caption{Image samples from the Brackish Dataset~\cite{pedersen2019detection}. In a majority of the sequences containing the \textit{small fish} class, there are multiple specimens forming a school of fish.}
    \label{fig:brackish}
\end{figure}
In 2019 the Brackish Dataset~\cite{pedersen2019detection} was published.
Its purpose was to advance object detection in brackish waters. 
It has been popular in the community since it was the first underwater detection dataset captured in non-tropical waters.
The recordings of the dataset were captured nine meters below the surface in brackish waters and consist of 89 sequences in total.
The sequences contain manually annotated bounding boxes of six coarse classes: \textit{fish}, \textit{crab}, \textit{shrimp}, \textit{starfish}, \textit{small fish}, and \textit{jellyfish}. 
Examples from the original dataset can be seen in \figref{brackish}.

\subsection{Dataset Overview}
In this work, we propose to expand the Brackish Dataset to include a MOT task.
Therefore, we provide a new set of ground truth annotations for every sequence, based on the MOTChallenge annotation style~\cite{dendorfer2020mot20}.
Additionally, we present 9 new sequences focused on the \textit{small fish} class, which gives a total of 98 sequences for the MOT task of the Brackish Dataset which we name \textbf{BrackishMOT}.
The \textit{small fish} class is especially relevant for the MOT task as it contains species that exhibit social and schooling behavior as illustrated in \figref{brackishB}.
The ground truth files are comma-separated and include annotations per object in the following structure:
\begin{lstlisting}[basicstyle=\footnotesize]
<frame>, <id>, <left>, <top>, <width>, <height>,
<confidence>, <class>, <visibility>
\end{lstlisting}
where \texttt{left} and \texttt{top} are the x and y coordinates of the object's top-left corner of the bounding box.
Together with the \texttt{width} and \texttt{height} they describe the object's bounding box in pixels.
\texttt{confidence} and \texttt{visibility} are both set to 1 and 
an object keeps its \texttt{id} as long as it is within field of view.
There are rare cases where objects gets fully occluded in-frame and it is ambiguous to decide the ID of the object as it re-appears; in these cases, the object acquires a new ID.
The \texttt{class} is in the range 1-6 where: (1) \textit{fish}, (2) \textit{crab}, (3) \textit{shrimp}, (4) \textit{starfish}, (5) \textit{small fish}, and (6) \textit{jellyfish}.

\begin{figure}[]
    \centering
    \begin{subfigure}[t]{0.4\linewidth}
    \includegraphics[width=\linewidth]{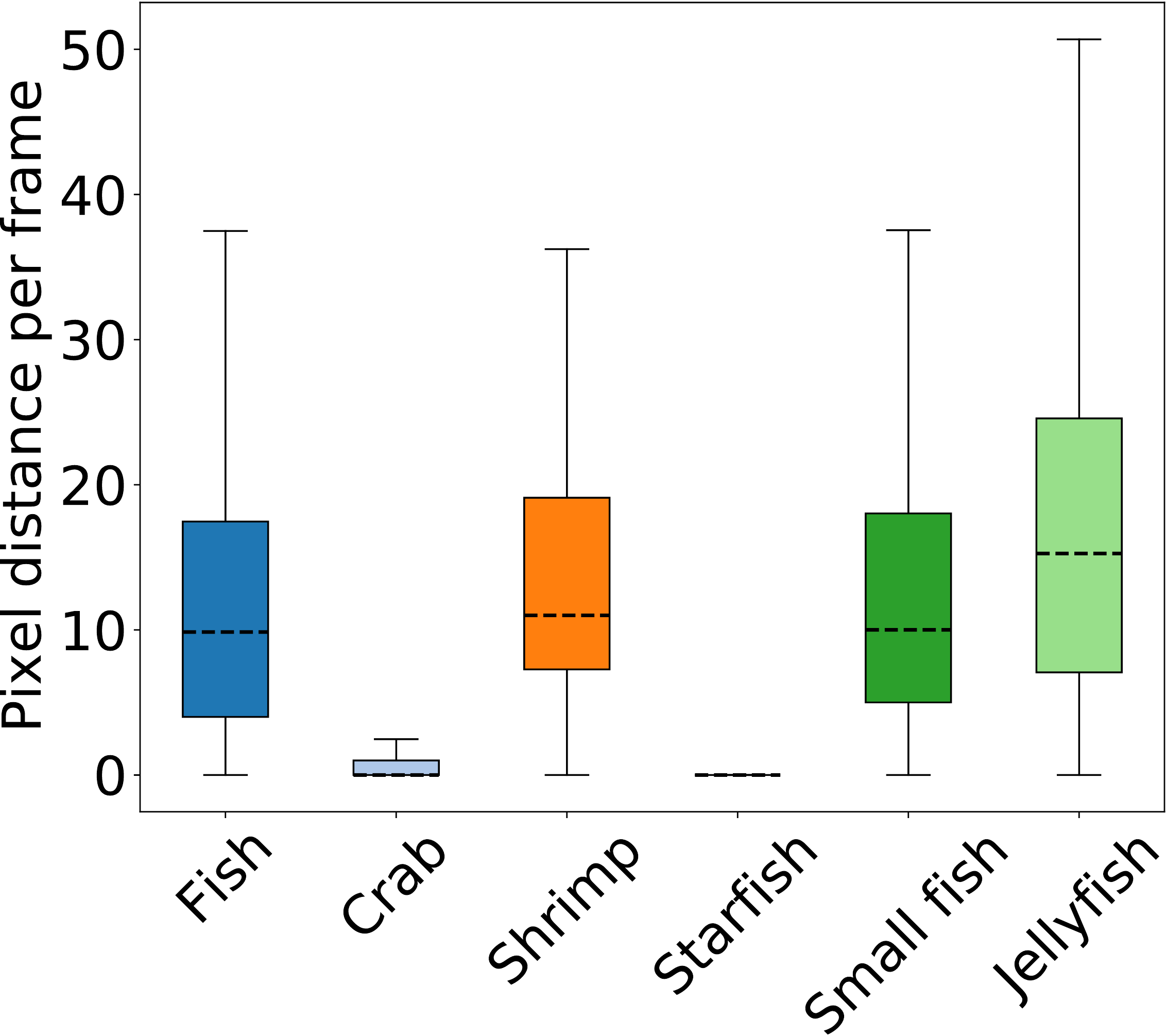} 
    \caption{Boxplot showing the distribution of the traveled distance between consecutive frames measured in pixels.}
    \label{fig:box}
    \end{subfigure}
    \qquad\qquad
    \begin{subfigure}[t]{0.4\linewidth}
        \centering
    \includegraphics[width=\linewidth]{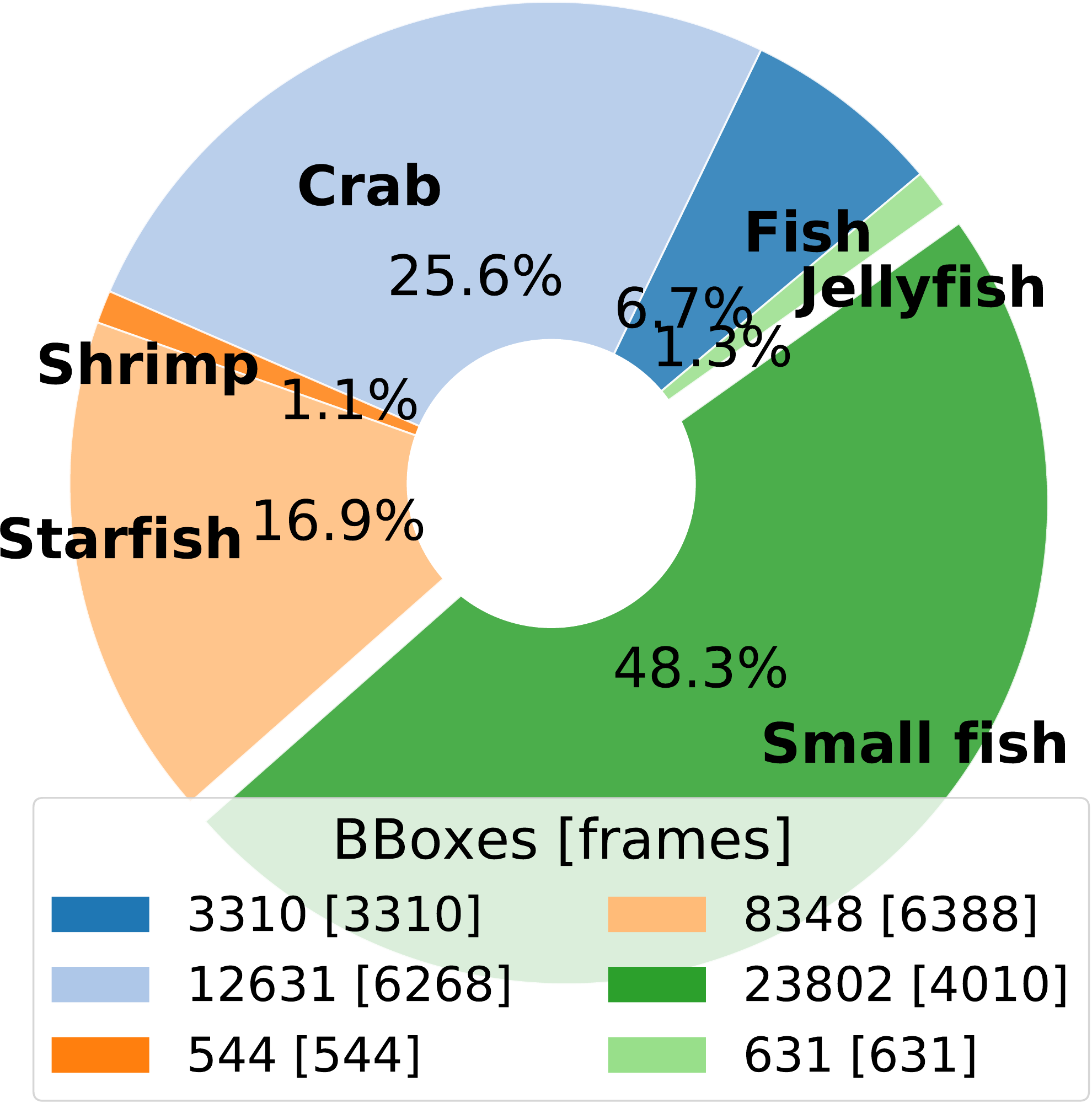}
    \caption{Class distribution of the brackishMOT dataset based on the number of bounding boxes.}
    \label{fig:class_all}
    \end{subfigure}
    \caption{Plots describing the composition of the brackishMOT dataset with respect to motion and class distribution. For both plots, the data is from all the sequences.}
    \label{fig:box_and_pie}
\end{figure}

In \figref{box_and_pie} we present two charts illustrating the motion and class distribution for the dataset. 
We see that the \textit{crab} and \textit{starfish} classes barely move compared to the rest.
In addition to that, they are well-camouflaged and most often move along the seabed.
This constitutes a specific task as they are both hard to detect visually and from motion cues.
The class distribution presented in \figref{class_all} shows that the dataset is imbalanced with few occurrences of the \textit{shrimp}, \textit{fish}, and \textit{jellyfish} classes.
Furthermore, as the number of bounding boxes and frame occurrences are equal we can decipher that these three classes occur in the sequences as single objects.
As the \textit{small fish} class is the only class that exhibits erratic motion and appears in groups it is deemed the most interesting class with respect to MOT. 

%% file: sections/split.tex
\subsection{BrackishMOT Splits}


Creating balanced training and testing splits is important to ensure a fair evaluation of the tracker performance and to give an accurate depiction of the task.
To a large degree, this is a problem that has been overlooked in the creation of most MOT datasets due to the lack of a suitable metric.
This has changed with the recent introduction of the MOTCOM framework~\cite{pedersen2022motcom}.
MOTCOM is a metric that can estimate the complexity of MOT sequences based on the ground truth annotations and lay the foundation for creating more balanced data splits.
The metric is a combination of three sub-metrics that describe the level of occlusion (OCOM), non-linear motion (MCOM), and visual similarity (VCOM) for every sequence.
MOTCOM and the sub-metrics are all in the interval from 0 to 1 where a higher MOTCOM score means a more complex problem.
We aim to create splits that are approximately evenly complex.

\begin{figure}[h]
    \centering
    \begin{subfigure}[t]{0.48\textwidth}
    \includegraphics[width=\linewidth]{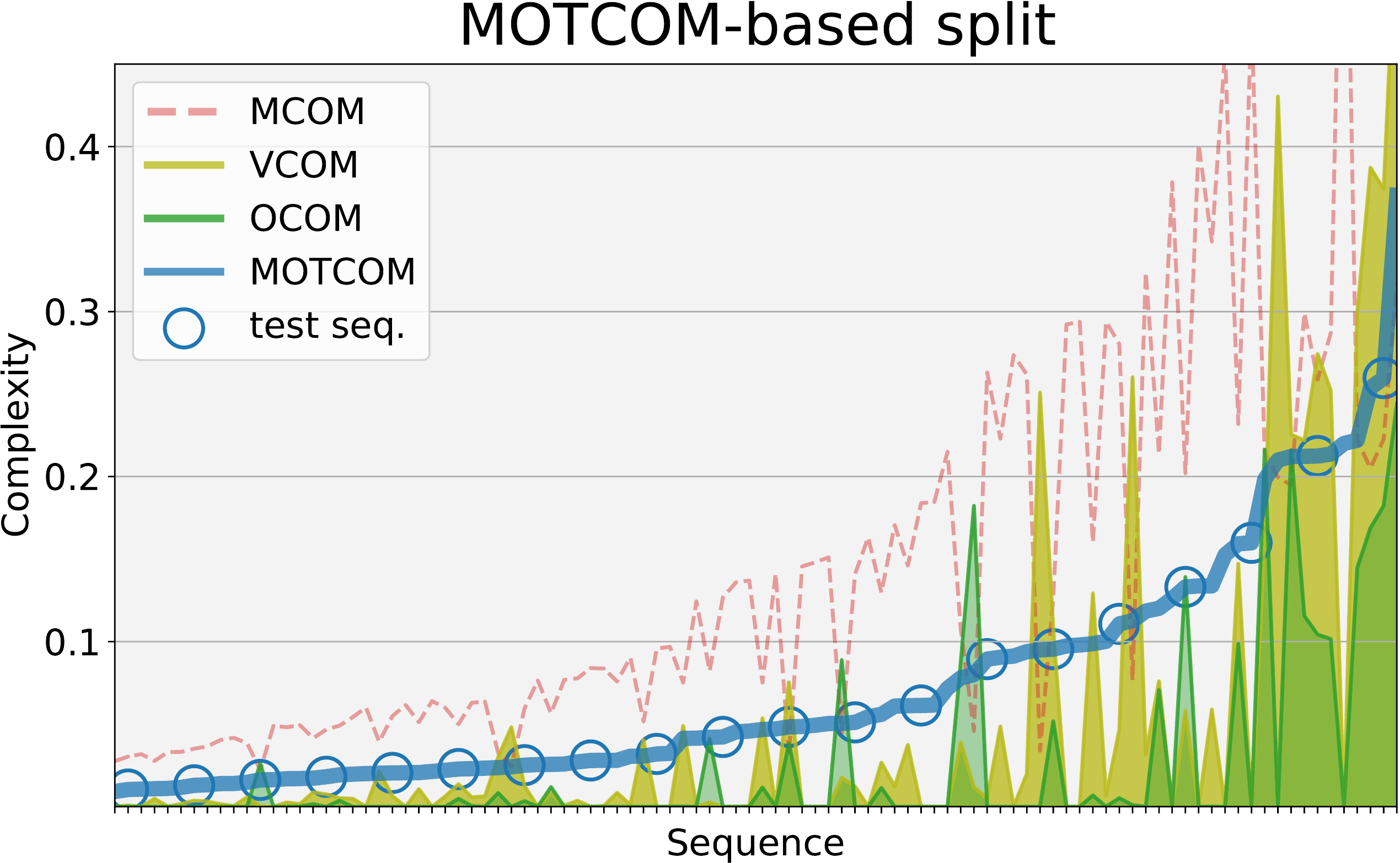}
    \caption{Sorting the sequences based on MOTCOM and taking every fifth to be included in the test split. This is the approach we follow.}
    \label{fig:motcom}
    \end{subfigure}
    \quad 
    \begin{subfigure}[t]{0.48\textwidth}
    \includegraphics[width=\linewidth]{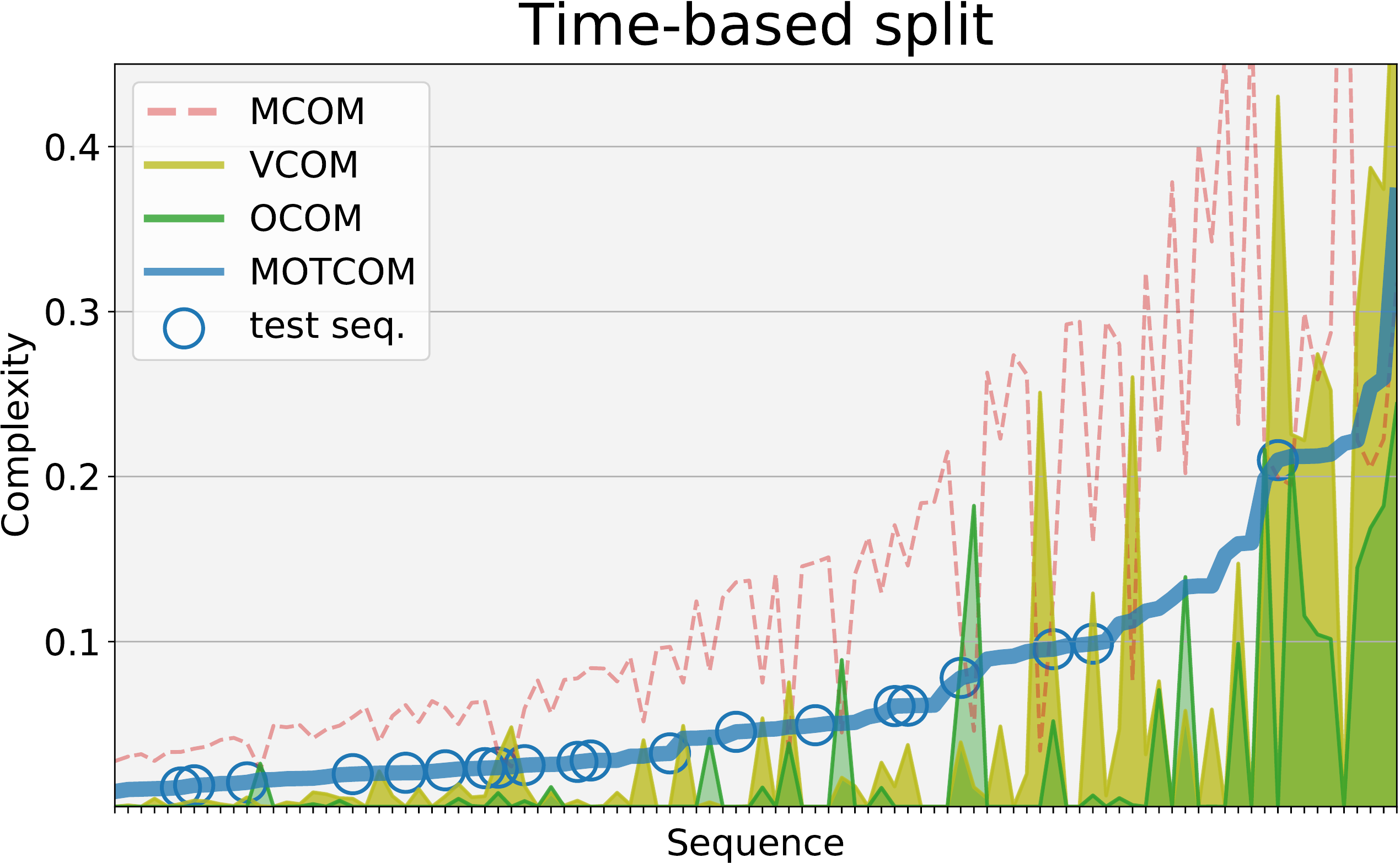}
    \caption{A typical test split consisting of the 20 first recorded sequences. This approach clearly skews the splits with respect to complexity.}
    \label{fig:motcom_split_timing}
   \end{subfigure} 
   \caption{These plots illustrate MOTCOM and the sub-metrics for all the BrackishMOT sequences. In both plots, the sequences are sorted based on their MOTCOM score. The circles mark the test sequences with respect to the split-scheme.}
   \label{fig:motcom_both}
\end{figure}

In \figref{motcom} we present MOTCOM and the sub-metrics for each sequence of the BrackishMOT dataset.
The metrics are calculated on basis of all six classes combined.
We see that the motion varies a lot between the individual sequences.
However, even though the motion is quite non-linear and complex for several sequences then both occlusion and visual similarity are very low.
This is due to the generally low number of objects in the scenes. 
A single jellyfish or shrimp may move fast and non-linearly, but if they are alone or in a scene with just a few objects they are less likely to be occluded or confused with other individuals. 

The sequences containing the \textit{small fish} class are generally exceptions to the above as they tend to score higher values in all three sub-metrics compared to the other sequences.
These sequences often include fish schools which means that they have a higher number of objects that moves more around and are more social compared to e.g., starfish and crabs on the seabed.
Therefore, the objects are more likely to be occluded and they are easier to confuse with each other as they look visually similar.

We create the splits based on the following scheme: we sort the sequences according to their MOTCOM score, then we pick the sequence with the highest MOTCOM score to be in the train split, the second highest goes into the test split, and from then on every fifth sequence goes into the test split while the rest goes into the train split.
This gives a total of 20 test sequences illustrated by the circles in \figref{motcom} and 78 train sequences.
If we, on the other hand, had chosen a typical scheme like picking the first 20 recorded sequences to be included in the test split and the rest in the train split, we would have had a significantly different dataset structure as illustrated in \figref{motcom_split_timing}.
With such a composition it is likely that trackers would generally perform better when evaluated on the test split, but it would be on false terms as the train split is significantly more challenging compared to the test split.
The opposite can of course also happen, but that is equally problematic.
For this reason, we make an informed split based on the MOTCOM scores.



To extend the proposed BrackishMOT dataset, we have developed a framework for creating synthetic underwater sequences based on phenomena observed in the BrackishMOT data as we believe that synthetic data is critical as a means to scale the availability of underwater training data.
The framework is described in the next section.



%% file: sections/syntheticData.tex
The proposed synthetic framework is built within the Unity game engine~\cite{unity}, using the built-in rendering pipeline and it is based on three main components: providing realistic fish meshes, modeling fish behavior, and building a realistically looking underwater environment. 
We provide options for each of the components in order to create a synthetic environment that resembles the BrackishMOT data, however, the proposed framework is easily extendable with other species, behavior models, and surroundings.
We will describe each of the components in the following sections.

\subsubsection{Fish Model}
\begin{figure}[]
     \centering
     \begin{subfigure}[b]{0.48\textwidth}
         \centering
         \includegraphics[width=\textwidth]{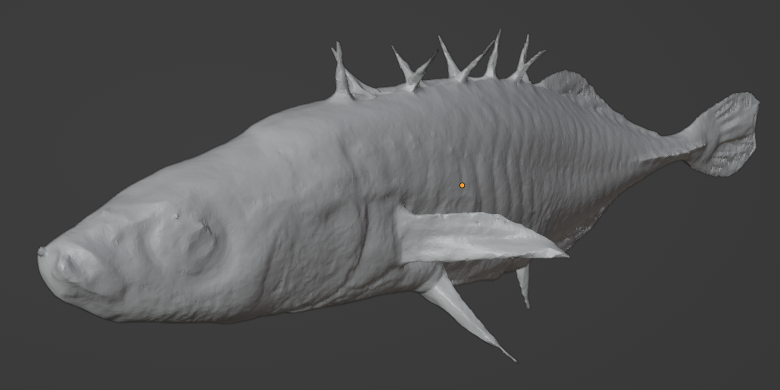}
         \caption{High-resolution mesh.}
         \label{fig:fishMesh_hd}
     \end{subfigure}
     \begin{subfigure}[b]{0.48\textwidth}
         \centering
         \includegraphics[width=\textwidth]{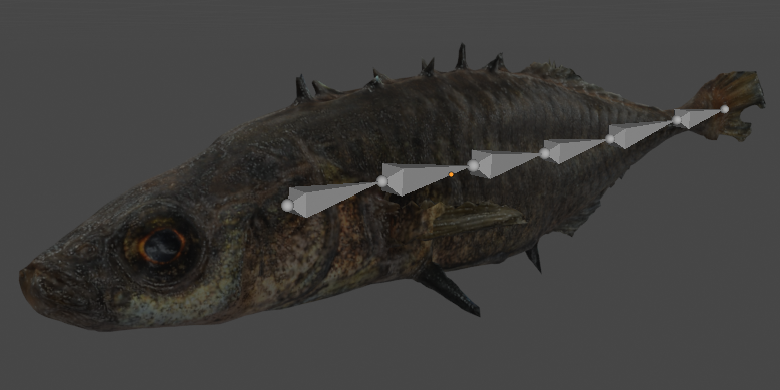}
         \caption{Low-resolution mesh with rig.}
         \label{fig:fishMesh_ld}
     \end{subfigure}
        \caption{Illustration of the \textit{stickleback} fish model used in our framework. (a) Initial high-resolution model and (b) decimated and rigged model for Unity.}
        \label{fig:fishModel}
\end{figure}

An illustration of the fish model used in our framework can be seen in \figref{fishModel}.
The model was taken from the fish database of images and photogrammetry 3D reconstructions \cite{kano2013online} and was selected as it visually resembles a \textit{stickleback}, which is the family of the \textit{small fish} class that most often occurs in schools in the Brackish dataset sequences according to the authors~\cite{pedersen2019detection}.
The 3D input model, shown in \figref{fishMesh_hd}, was decimated to 11,000 vertices. 
In order to preserve finer details of the mesh, a normal map was created from the high-resolution texture.
The down-sampled mesh can be seen in \figref{fishMesh_ld}.
Lastly, the model was rigged using the bones system in Blender~\cite{blender} to allow for smooth animations of the body and tail.
%

The number of spawned fish in each sequence is randomly selected within a range between 4 and 50 to resemble the diversity of the BrackishMOT sequences.
The initial pose, scale, and appearance (texture albedo and glossiness) for each fish varies between the sequences. 
A table with all the randomized parameters and their respective ranges can be found in the supplementary material.

\subsubsection{Behavior Model}
To approximate realistic fish schooling behavior, we use a boid-based behavioral model inspired by the work of C.W. Reynolds~\cite{reynolds1987flocks} and C. Hartman and B. Benes~\cite{hartman2006autonomous}.
Each fish considers the position and heading of all other fish in its neighborhood.
For each fish the velocity and heading is dependent on four factors: separation $\vec{s}$, cohesion $\vec{k}$, alignment $\vec{m}$, and leader $\vec{l}$.
Separation ensures avoidance of collisions with other members of the school.
Cohesion is a force that drives the fish to seek the center of the neighborhood.
Alignment is the drive of individual fish to match the others' velocity.
Leader is a direction towards where a given leader is heading and for each fish, the leader is the neighbor with a heading vector closest to the fish's own heading vector.
The steering vector is given by
\begin{equation}\label{eq:steering}
    \vec{steer} = S\vec{s} + K\vec{k} + M\vec{m} + L\vec{l},
\end{equation}
where $S$, $K$, $M$, and $L$ are weights for the separation, cohesion, alignment, and leader forces, respectively. 
A more detailed description of the behavior model can be found in the supplementary material.

\begin{figure}[]
     \centering
     \begin{subfigure}[b]{0.48\textwidth}
         \centering
         \includegraphics[width=\textwidth]{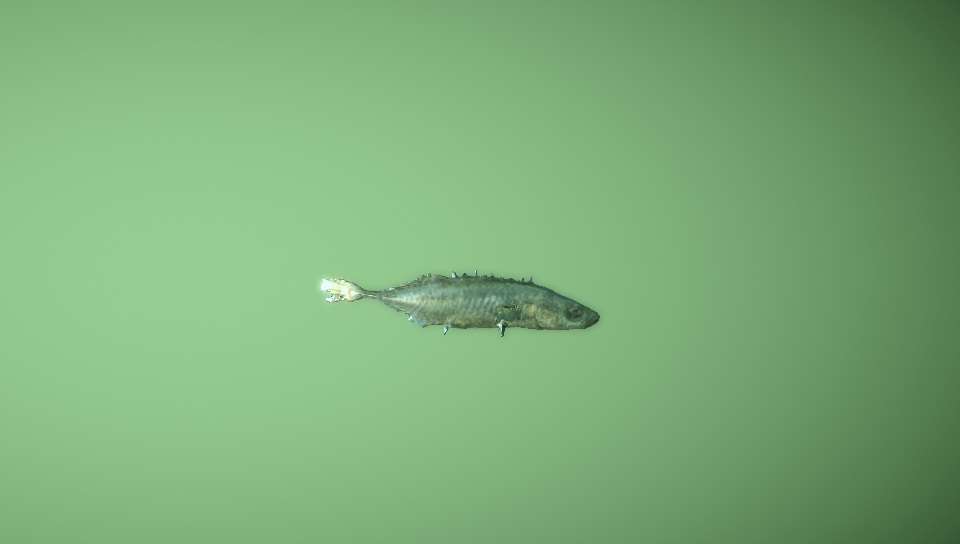}
         \caption{\textbf{Synth}}
         \label{fig:synthEnvBasis}
     \end{subfigure}
     \begin{subfigure}[b]{0.48\textwidth}
         \centering
         \includegraphics[width=\textwidth]{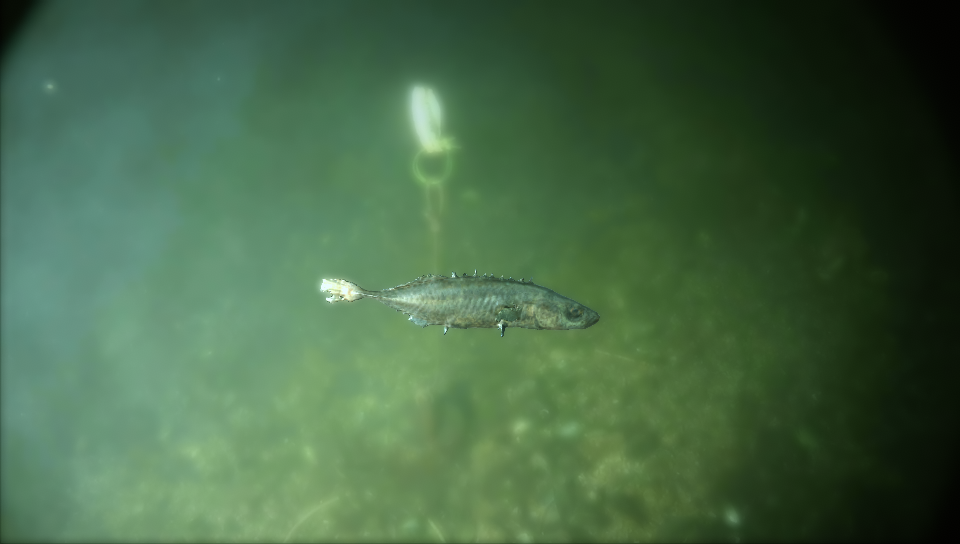}
         \caption{\textbf{Synth\textsubscript{B}}}
     \end{subfigure}
     \begin{subfigure}[b]{0.48\textwidth}
         \centering
         \includegraphics[width=\textwidth]{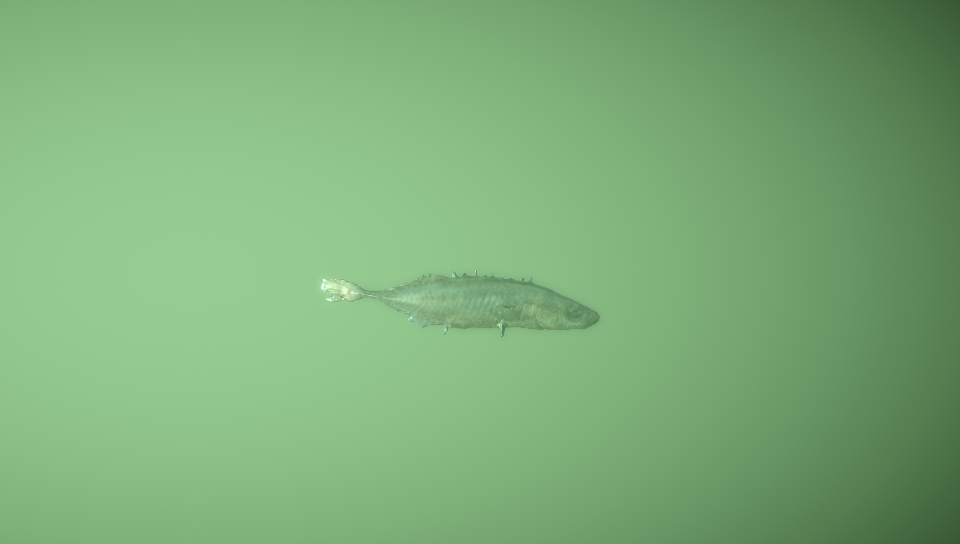}
         \caption{\textbf{Synth\textsubscript{T}}}
     \end{subfigure}
     \begin{subfigure}[b]{0.48\textwidth}
         \centering
         \includegraphics[width=\textwidth]{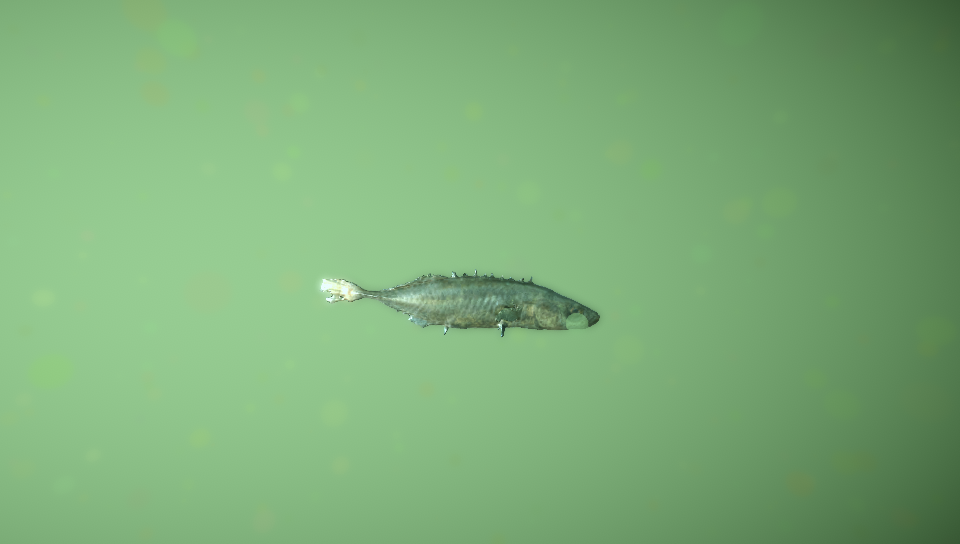}
         \caption{\textbf{Synth\textsubscript{D}}}
     \end{subfigure}
     \begin{subfigure}[b]{0.48\textwidth}
         \centering
         \includegraphics[width=\textwidth]{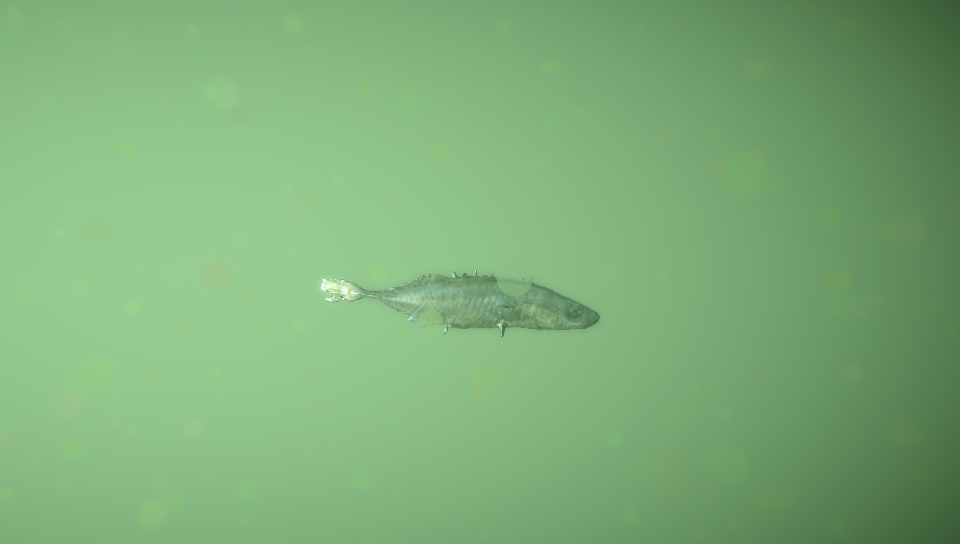}
         \caption{\textbf{Synth\textsubscript{TD}}}
     \end{subfigure}
     \begin{subfigure}[b]{0.48\textwidth}
         \centering
         \includegraphics[width=\textwidth]{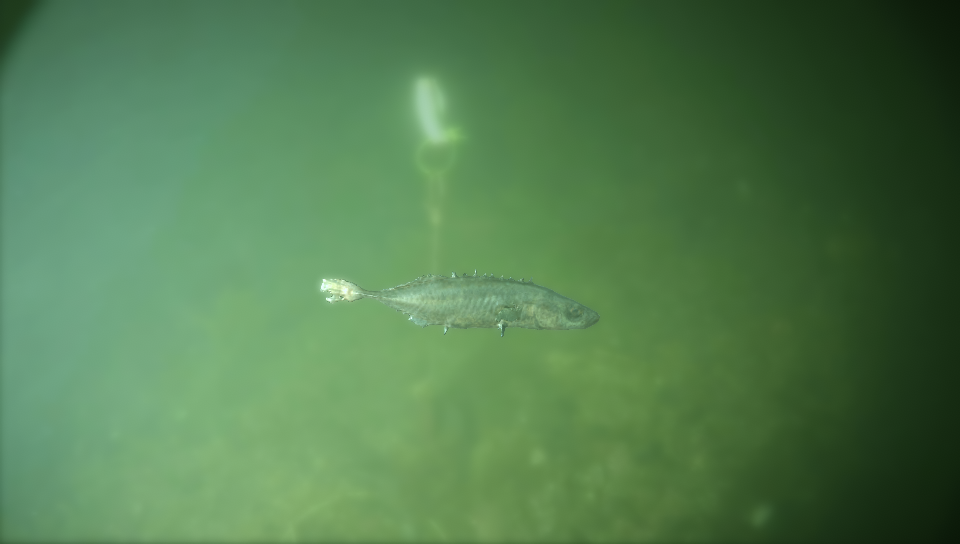}
         \caption{\textbf{Synth\textsubscript{BT}}}
     \end{subfigure}
     \begin{subfigure}[b]{0.48\textwidth}
         \centering
         \includegraphics[width=\textwidth]{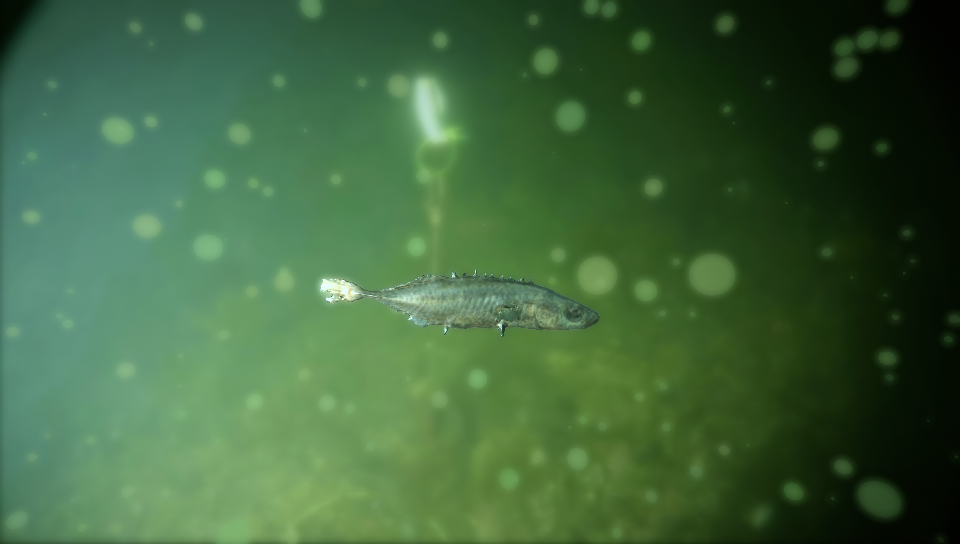}
         \caption{\textbf{Synth\textsubscript{BD}}}
     \end{subfigure}
     \begin{subfigure}[b]{0.48\textwidth}
         \centering
         \includegraphics[width=\textwidth]{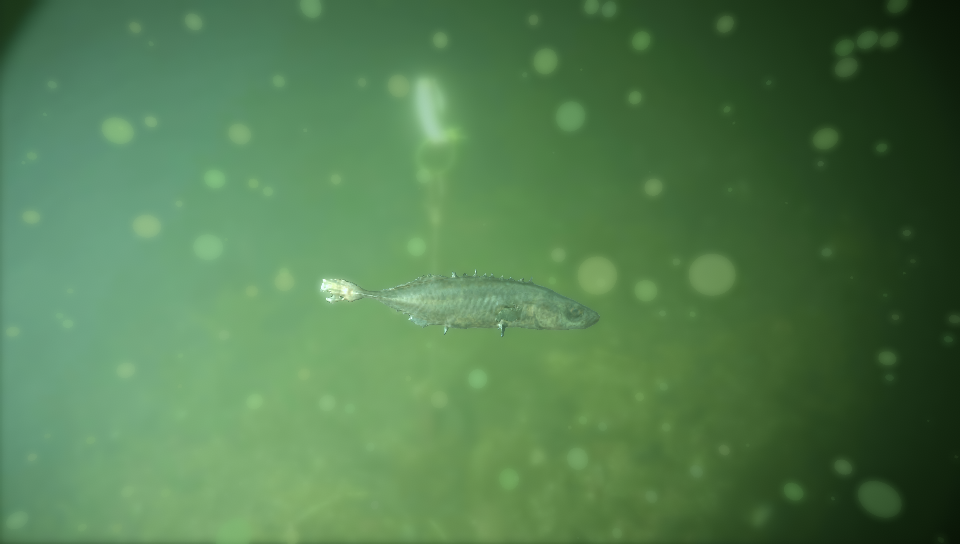}
         \caption{\textbf{Synth\textsubscript{BTD}}}
     \end{subfigure}
        \caption{Visualisation of different conditions of our synthetic environment. \textbf{(a)} Plain background, no turbidity, no distractors \textbf{(Synth)}. \textbf{(b)} Video background, no turbidity, no distractors \textbf{(Synth\textsubscript{B})}. \textbf{(c)} Plain background, turbidity, no distractors \textbf{(Synth\textsubscript{T})}. \textbf{(d)} Plain background, no turbidity, with distractors \textbf{(Synth\textsubscript{D})}. \textbf{(e)} Plain background, with  turbidityand distractors \textbf{(Synth\textsubscript{DT})}. \textbf{(f)} Video background with turbidity, but without distractors \textbf{(Synth\textsubscript{BT})}. \textbf{(g)} Video background with distractor, but no turbidity \textbf{(Synth\textsubscript{BD})}. \textbf{(h)} Video background with turbidity and distractors \textbf{(Synth\textsubscript{BTD})}.}
        \label{fig:synthEnv}
\end{figure}

\subsubsection{The Surrounding Environment}
To investigate how changes in the environment impact tracker performance, we design the synthetic environment based on three variables: turbidity, background, and distractors.

Turbidity represents tiny floating particles in the water that engulfs the scene like a fog that intensifies as the distance between the object and the camera increases.
The visibility varies to a large degree in the BrackishMOT sequences due to this phenomenon.
We implement the turbidity effect using a custom-made Unity material with adjustable transparency and post-processing effects of depth of field and color grading.
The color of the material spans between grey and green to resemble the turbidity observed in the BrackishMOT sequences.
Both the color and intensity vary between the generated sequences.

We use videos from the Brackish dataset without fish as the background to make the scene more realistic.
We include a range of background sequences and augment them by saturation, color, and blur to increase variation.
When no background is present, we use a monotone color that matches the color of the turbidity.

Lastly, we introduce distractors~\cite{tobin2017domain,tremblay2018training}, which represent floating particles.
The BrackishMOT sequences have been captured in shallow water where the current is often strong.
The combination of strong current and shallow water induces floating plant material and resuspended sediments often occur in front of the camera as unclear circular bodies.
To simulate this phenomenon we implement distractors as spheres with varying scales, levels of transparency, and color.
The color range spans between grey and green as with the turbidity and monotone background.
The number of distractors vary between the sequences and each distractor is spawned in a random position and is randomly moved to a new position between each frame.

Each of the environmental variables adds a layer of complexity to the synthetic scene based on phenomena observed in the real sequences.
Combinatorial variations of the variables give us eight synthetic environments, which can be seen in \figref{synthEnv}.
Each generated video sequence is 10 seconds long and contains 150 frames animated with a frame rate of 15 FPS, which resembles the sequences of the BrackishMOT dataset.
We include 50 sequences for each environment variation.
The synthetic framework is general as all parameters can be adjusted to fit other underwater environments, e.g., using another video background would significantly alter the visuals of the sequences, or one could change the current or add new models to the scene.
Source code and guides to using the framework can be found on the project page \textit{(URL upon paper acceptance)}.

%% file: sections/results_new.tex
It is notoriously difficult for humans to visually track fish of the same species in video sequences captured in the wild. 
This is especially true when the water is turbid, the camera resolution is low, and the objects swim close to each other as is the case in some of the BrackishMOT sequences.
This indicates that visual cues for re-identifying the objects are not pronounced and likely not reliable for solving this specific problem.
Therefore, we conduct experiments based on the state-of-the-art tracker CenterTrack~\cite{zhou2020tracking}, which tracks objects as points and focuses on associating objects locally between consecutive frames with little emphasis on visual features.

As a basis for our experiments, we use two pre-trained models provided by the authors of CenterTrack and fine-tune on top of them to reduce training time and minimize the potential of overfitting. 
We name the base models CT-COCO and CT-ImNet, where
CT-COCO has been pre-trained on the MS COCO dataset~\cite{lin2014microsoft} and CT-ImNet has been pre-trained on the ImageNet dataset~\cite{deng2009imagenet}.
Both models have a similar architecture with a DLA-34~\cite{yu2018deep} backbone.
We train all our models with a batch size of 12 and a learning rate of 1.25e-4 and we resize and pad the BrackishMOT images from 1920x1080 to 960x544 following the strategy proposed by the CenterTrack authors.

First, we evaluate how pre-training on the two large-scale datasets MS COCO and ImageNet affects CenterTrack's ability to learn from the BrackishMOT sequences.
We then extend these results by introducing training strategies for including synthetic sequences, to investigate the potential benefits of using synthetic tracking data in underwater environments.
We evaluate all our models based on conventional MOT performance metrics like the Multiple Object Tracking Accuracy metric (MOTA) from the CLEAR MOT metrics~\cite{bernardin2008evaluating}, the ID F1 score (IDF1)~\cite{ristani2016performance}, and the recent Higher Order Tracking Accuracy metric (HOTA)~\cite{luiten2020hota}.

\subsubsection{Training on Real Data}
We fine-tune the base models on the BrackishMOT train split, which consists of 78 sequences, for 30 epochs and name the new models CT-COCO-Brack and CT-ImNet-Brack.
Evaluating these models on the 20 sequences of the BrackishMOT test split gives us an indication of the performance to be expected from fine-tuning a state-of-the-art tracker on manually annotated real data.
The HOTA, MOTA, and IDF1 results are presented in \tableref{quantitative_real_data} along with detections (\textit{Dets}), ground truth detections (\textit{GT dets}), IDs, ground truth IDs (\textit{GT
IDs}), and ID switches (\textit{IDSW}).

We see that both models deliver promising results although the model pre-trained on ImageNet outperforms the model pre-trained on MS COCO.
This indicates that ImageNet is better suited as a foundation for detecting and tracking objects in this type of underwater environment. 
We will investigate whether this is also the case when including synthetic data in the following sections.

\begin{table}[t]
\centering
\caption{Performance of the CenterTrack models fine-tuned on the BrackishMOT train sequences and evaluated on the BrackishMOT test split.}
\label{tab:quantitative_real_data}
\begin{tabular*}{\textwidth}{@{\extracolsep{\fill}}lccccccccc@{}}
\toprule
Model & HOTA$\uparrow$ & MOTA$\uparrow$& IDF1$\uparrow$ & Dets & GT dets & IDs & GT IDs & IDSW \\ \midrule
CT-COCO-Brack          & 0.36 & 0.37 & 0.39 & 10270 & 14670 & 887
 & 182 & 493 \\
CT-ImNet-Brack         & \textbf{0.38} & \textbf{0.43} & \textbf{0.44} & 10056 & 14670 & 755 & 182 & 464 \\ \bottomrule
\end{tabular*}
\end{table}

\subsubsection{Training on Synthetic Data}
Next, we investigate whether the base models can be taught to track fish in real sequences if they are fine-tuned strictly on synthetic data. 
We do this by studying how the combinations of the environment with turbidity (T), background (B), and distractors (D) affect the tracking performance.
We fine-tune the base models for 10 epochs on the eight different sets of synthetic sequences.
The synthetic sequences only contain the \textit{small fish} class, therefore, we evaluate the fine-tuned models on a sub-set of the BrackishMOT test split consisting of the sequences with the \textit{small fish} class.
We name this sub-set the 'small fish split' and it contains eight sequences (the list of sequences is presented in \tableref{qualitative} as part of another evaluation).

\begin{table}[]
\centering
\caption{Performance of CenterTrack models trained strictly on variations of the synthetic dataset. The models have been evaluated on the small fish split.}
\label{tab:synth_datasets}
\begin{tabular*}{\textwidth}{@{\extracolsep{\fill}}lccc c lccc@{}}
\toprule
\textbf{CT-COCO} & HOTA$\uparrow$ & MOTA$\uparrow$& IDF1$\uparrow$ & & \textbf{CT-ImNet} & HOTA$\uparrow$ & MOTA$\uparrow$& IDF1$\uparrow$ \\ \midrule
Synth        & 0.08 & -0.17 & 0.07 &\qquad\qquad &  Synth & 0.08 & -0.90 & 0.06  \\
Synth$_B$  & 0.12 & -0.21 & 0.12 && Synth$_{B}$     & 0.14 & 0.05 & 0.17  \\
Synth$_T$      & 0.08 & -2.02 & 0.06 &&  Synth$_{T}$     & 0.06 & -0.93 & 0.04                         \\
Synth$_D$     & 0.09 & -0.12 & 0.09  &&  Synth$_{D}$     & 0.08 & 0.03 & 0.08                          \\
Synth$_{TD}$    & 0.12 & -0.14 & 0.12&&  Synth$_{TD}$    & 0.06 & 0.02 & 0.05                          \\
Synth$_{BT}$   & 0.19 & 0.16 & 0.21  &&  Synth$_{BT}$    & \textbf{0.19} & -0.24 & \textbf{0.21}      \\
Synth$_{BD}$  & 0.15 & 0.08 & 0.16   &&  Synth$_{BD}$    & 0.17 & \textbf{0.13} & 0.19                 \\
Synth$_{BTD}$ & \textbf{0.21} & \textbf{0.18} & \textbf{0.24}   &&  Synth$_{BTD}$   & 0.13 & 0.00 & 0.14  \\ \midrule\midrule
CT-COCO-Brack & 0.37 & 0.47 & 0.43  &&  CT-ImNet-Brack   & 0.39 & 0.50 & 0.46 \\ \midrule
\end{tabular*}
\end{table}

The results of the synthetically trained models evaluated on the small fish split are presented in \tableref{synth_datasets} along with the results of the CT-COCO-Brack and CT-ImNet-Brack models for comparison. 
Although the synthetic models only perform up to half as well as the models trained on real data, we see that it is in fact possible to train CenterTrack to be able to detect and track the \textit{small fish} class without ever seeing real images of the class.
The feature that seems to increase the tracking performance the most is by adding background videos whereas the turbidity and distractors give mixed results.
We see that the CT-COCO model performs the best when fine-tuned on the Synth$_{BTD}$ sequences while it is more unclear what benefits the CT-ImNet model the most, however, a good compromise seems to be the Synth$_{BD}$ sequences.

\subsubsection{Two Strategies for Training on both Synthetic and Real Data}
Previously, we found the synthetic sequences best suited for teaching the base models to track the \textit{small fish} class.
Now, we examine whether the CT-COCO model fine-tuned on Synth$_{BTD}$ and the CT-ImNet model fine-tuned on Synth$_{BD}$ provide better foundations compared to the base models.
We fine-tune on top of these models for 30 epochs on the BrackishMOT train sequences and name these two-step fine-tuned models CT-COCO-Synth$_{BTD}$ and CT-ImNet-Synth$_{BD}$.
Additionally, we examine the potential benefits of combining real and synthetic data in a single training step by fine-tuning the base models for 30 epochs on a combination of the BrackishMOT train and Synth$_{BTD}$ sequences.
We name these the CT-COCO-Mix and CT-ImNet-Mix models.

\begin{table}[b]
\centering
\caption{Baseline tracking results for the BrackishMOT test split and the small fish split.}
\label{tab:quantitative}
\begin{tabular*}{\textwidth}{@{\extracolsep{\fill}}lclccc clccc@{}}
\toprule
Model                  &\qquad\qquad&& HOTA$\uparrow$ & MOTA$\uparrow$& IDF1$\uparrow$ &\quad\qquad&&  HOTA$\uparrow$ & MOTA$\uparrow$& IDF1$\uparrow$\\ \midrule
CT-COCO-Brack && \parbox[t]{5mm}{\multirow{6}{*}{\rotatebox[origin=c]{90}{\textbf{Test split}}}} & 0.36 & 0.37 & 0.39 && \parbox[t]{5mm}{\multirow{6}{*}{\rotatebox[origin=c]{90}{\textbf{Small fish split}}}}& 0.37 & 0.47 & 0.43 \\
CT-COCO-Synth$_{BTD}$  &&& 0.36 & 0.38 & 0.39 &&& 0.39 & 0.47 & 0.44 \\
CT-COCO-Mix            &&& 0.36 & 0.37 & 0.39 &&& 0.37 & 0.46 & 0.43 \\
CT-ImNet-Brack         &&& 0.38 & 0.43 & 0.44 &&& 0.39 & 0.50 & 0.46 \\
CT-ImNet-Synth$_{BT}$  &&& 0.38 & 0.42 & 0.41 &&& \textbf{0.41} & \textbf{0.52} & 0.48 \\
CT-ImNet-Mix           &&& \textbf{0.40} & \textbf{0.44} & \textbf{0.45} &&& \textbf{0.41} & \textbf{0.52} & \textbf{0.49} \\ \bottomrule
\end{tabular*}
\end{table}

Baseline results for the models are presented in \tableref{quantitative} for both the regular test split and the small fish split.
We use the small fish split to examine whether the models trained on the synthetic data overfits to the \textit{small fish} class.
Generally, we see a tendency that the ImageNet pre-trained models perform better than the models pre-trained on the MS COCO dataset, which indicates that the ImageNet dataset lays a stronger foundation for detecting the objects of the BrackishMOT dataset.
Furthermore, the CT-COCO models do not seem to benefit from the synthetic data, which is in contrast to the results presented in \tableref{synth_datasets} that showed that fine-tuning the CT-COCO model on the Synth$_{BTD}$ sequences gave the best performing purely synthetically trained tracker.

For the CT-ImNet-Synth$_{BT}$ model we see a slight decrease in MOTA and IDF1 when evaluating on the test split, but an increase in the \textit{small fish} sequences, this indicates that the model learns from the synthetic data to better track the \textit{small fish} objects but at the expense of some of the other classes.
The CT-ImNet-Mix model exhibits similar performance as the CT-ImNet-Synth$_{BT}$ model on the \textit{small fish} sequences.
However, the performance is also increased when looking at all the test sequences, which indicates that the ability to track the other classes is maintained using this training strategy.

\subsection{Qualitative Evaluation}
In the previous evaluation, we found the overall best-performing model to be CT-ImNet-Mix.
In this section, we analyse how the model performs on each of the eight sequences from the small fish split.
The qualitative results of the CT-ImNet-Mix model when evaluated on the small fish split are presented in \tableref{qualitative}.
When we inspect the brackishMOT-93 and brackishMOT-95 sequences we see that they have 45 and 1 GT IDs, respectively. 
However, both sequences score a HOTA performance of 0.44 indicating that a higher number of objects does not seem to have a significantly negative impact on the tracking performance.
If we look at BrackishMOT-67 it has four GT IDs but the tracker only manages to get a HOTA score of 0.18.
Inspecting the BrackishMOT-67 sequence visually shows that it contains a single medium-sized object of the \textit{small fish} class, which the model tracks well throughout the sequence, however, there are also three tiny objects of the \textit{small fish} class near the seabed that the model largely fails to detect and this penalizes the tracking performance greatly.

\begin{table}[]
\centering
\caption{Performance of CT-ImNet-Mix model on the sequences of the small fish split.}
\label{tab:qualitative}
\begin{tabular*}{\textwidth}{@{\extracolsep{\fill}}lccccccccc@{}}
\toprule
Sequence        & HOTA$\uparrow$ & MOTA$\uparrow$& IDF1$\uparrow$ & Dets & GT dets & IDs & GT IDs & IDSW \\ \midrule
brackishMOT-50  & 0.40 & 0.46 & 0.47 & 1785 & 2129 & 105 &  17 & 50   \\
brackishMOT-55  & 0.35 & 0.46 & 0.43 & 2318 & 3192 & 187 &  37 & 112  \\
brackishMOT-56  & 0.53 & 0.41 & 0.75 & 80   & 87   & 7   &  1  & 4    \\
brackishMOT-67  & 0.18 & 0.11 & 0.10 & 173  & 636  & 31  &  4  & 7    \\
brackishMOT-90  & 0.61 & 0.53 & 0.74 & 401  & 426  & 18  &  3  & 7    \\
brackishMOT-93  & 0.44 & 0.51 & 0.51 & 1450 & 1567 & 160 &  45 & 82   \\ 
brackishMOT-95  & 0.44 & 0.38 & 0.39 & 619  & 148  & 28  &  1  & 0    \\ 
brackishMOT-98  & 0.49 & 0.58 & 0.60 & 1728 & 1930 & 137 &  36 & 80   \\ \bottomrule
\end{tabular*}
\end{table}

%

%% file: sections/conclusion.tex
We propose a new underwater multi-object tracking dataset named BrackishMOT, which is an extension of the Brackish dataset captured in turbid waters in Denmark.
This is the first and only dataset of its kind and it is a necessary step towards increasing the capability of underwater trackers as there currently only exist very few underwater tracking datasets and they have all been captured in clear tropical waters.
Furthermore, we propose a framework for generating synthetic underwater MOT sequences and present baseline results based on fine-tuning CenterTrack using three different training strategies.
We show that tracking performance can be increased by including sequences generated by the proposed synthetic framework in the training procedure.